\documentclass[10pt,twocolumn]{article}
\usepackage[letterpaper,top=0.75in,bottom=0.76in,left=0.73in,right=0.73in]{geometry}
\usepackage[T1]{fontenc}
\usepackage{times}
\usepackage{microtype}
\usepackage{graphicx}
\usepackage{amsmath,amssymb,amsthm}
\usepackage{booktabs,multirow,tabularx,makecell}
\usepackage{url,xcolor}
\usepackage{fancyhdr}
\fancypagestyle{plain}{\fancyhf{}\fancyhead[R]{\scriptsize\thepage}}
\usepackage[font=small,labelfont=bf]{caption}
\usepackage{enumitem}
\usepackage{placeins}
\usepackage[breaklinks=true,hidelinks]{hyperref}
\setlist{nosep,leftmargin=1.3em}
\newcommand{\model}{ProtoBridge-CLIP}
\newcommand{\Lglobal}{\mathcal{L}_{\mathrm{global}}}
\newcommand{\Lconcept}{\mathcal{L}_{\mathrm{finding}}}
\newcommand{\Lrel}{\mathcal{L}_{\mathrm{relation}}}
\newcommand{\KL}{D_{\mathrm{KL}}}

\DeclareMathOperator{\sigmoid}{sigmoid}
\newtheorem{proposition}{Proposition}
\title{\vspace{-1.25em}\bfseries ProtoBridge-CLIP: Observation-Masked Finding Prototypes for Medical Image--Report Representation Learning}
\author{Jinrui Wu$^{1}$ and Yun Peng$^{1}$\\[0.3em]\normalsize $^{1}$Fudan University, Shanghai, China}
\date{}
\begin{document}
\twocolumn[\maketitle\vspace{-1.9em}
\begin{abstract}
Radiology reports distinguish affirmed findings, explicit negations, and unmentioned concepts, but conventional global image--report contrastive learning often compresses these different evidence states into a single study-level target. We introduce \model{}, which combines symmetric global image--report alignment with observation-masked finding supervision and a soft relation objective over affirmed report concepts. Text-derived finding prototypes are aligned with local visual tokens, while uncertain and unmentioned concepts are excluded from direct finding labels. We evaluate the framework on three CT benchmarks and five chest-radiograph benchmarks, together with component ablations, paired and clinical retrieval, localization, perturbation robustness, calibration, and per-finding analyses. \model{} achieves macro-AUROC values of 73.6\%, 66.5\%, and 68.8\% on CT-RATE, RAD-ChestCT, and INSPECT, and 80.2\%, 91.4\%, 78.5\%, 86.5\%, and 90.8\% on ChestXray14, CheXpert, PadChest, RSNA, and SIIM-ACR, respectively. Relative to the matched global-only controls, the findings and relation objectives yield consistent gains across the evaluated datasets. These results demonstrate the practical value of treating observed and unobserved report concepts differently during medical image--text representation learning.
\end{abstract}\vspace{0.45em}]

\section{Introduction}
Large-scale paired image--text pretraining has made visual recognition possible with natural-language prompts \cite{radford2021learning,jia2021scaling}. Radiology offers a related but distinctly clinical setting: a single radiograph or CT study can be accompanied by a narrative of imaging findings, uncertainty, comparisons, and recommendations. Global image--report alignment has established a strong foundation \cite{zhang2022contrastive,huang2021gloria,boecking2022making,tiu2022expert}, and later systems have incorporated explicit medical language, structured knowledge, and more expressive interactions \cite{zhang2023knowledge,wu2023medklip,lai2024carzero,luo2025devide}. Volumetric CT pretraining now offers a natural test of whether these relations survive the move from two-dimensional imaging to large three-dimensional fields of view \cite{hamamci2026generalist,lai2025brgsa}.

A medical report, however, is not an exhaustive list of all possible clinical observations. A phrase such as ``small right upper-lobe nodule'' supplies positive evidence for a nodule; ``no pleural effusion'' supplies evidence for an explicit negative; failure to discuss emphysema is usually not equivalent to a negative diagnosis. The distinction matters when a report is repurposed as weak supervision: the absence of a mention must not automatically create a negative target. Clinical relation extraction systems, structured report datasets, and uncertainty-aware benchmark practice provide relevant precedents \cite{jain2021radgraph,irvin2019chexpert,bustos2020padchest}.

Global contrastive learning also compresses several findings into one report embedding. Two studies with the same nodule diagnosis may differ in age, anatomy, modality, report length, and unrelated pathology. Standard batch negatives do not explicitly encourage finding-level consistency among such studies. Conversely, forcing all reports with shared labels into a single class can erase clinically relevant variability. This suggests an intermediate form of supervision: align each affirmed or negated statement to candidate local visual evidence while retaining a graded description of patient similarity.

The proposed \model{} formulation makes that interface explicit. A bank of text-derived finding prototypes is compared against the local tokens of the image encoder, and report statements control a masked finding-level loss. A second loss translates similarity between affirmed report concepts into a soft distribution over other examinations. The original symmetric global contrastive term remains in place to preserve general image--text correspondence. CT and chest-radiograph (CXR) branches use modality-specific encoders with a shared conceptual supervision scheme.

A central methodological question is how a learning system should handle the incompleteness of narrative supervision. A report may be clinically accurate while leaving many visually assessable conditions unmentioned. Treating every omitted concept as absent makes the supervision depend on reporting habits rather than only on image evidence. Observation masking does not resolve all label noise, but it makes one important assumption explicit and testable: only assertions extracted with a defined state are assigned direct positive or negative targets. Because the same extracted statements also enter relational supervision, errors in assertion parsing can propagate through two pathways, which motivates a dedicated extraction audit and targeted ablations.

The evaluation is organized around distinct types of evidence. Diagnostic AUROC addresses discrimination; paired retrieval measures study-level image--report correspondence; label-aware retrieval addresses agreement in the abnormalities represented by retrieved cases; and spatial grounding asks whether finding scores depend on image locations consistent with independent annotations. Corruption testing and calibration describe further operational properties. These endpoints are complementary and should not be collapsed into a single claim about clinical utility, particularly when models are trained using different modalities or pretraining corpora.

\begin{figure*}[t]
 \centering
 \includegraphics[width=.97\textwidth]{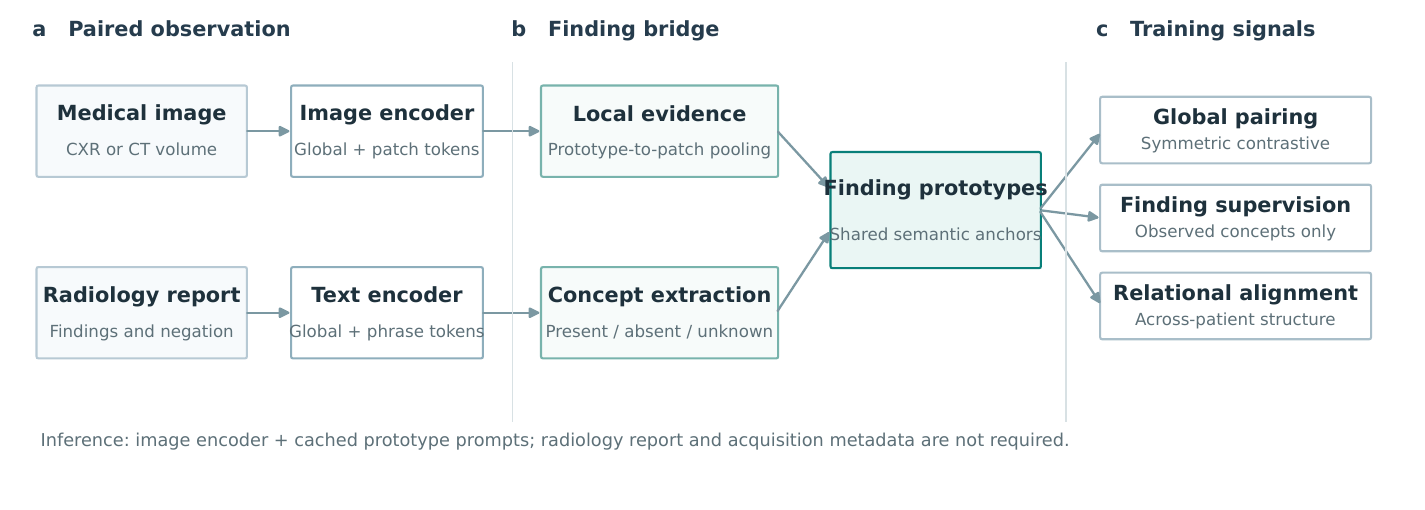}
 \caption{\textbf{Training organization of \model{}.} Global image--report alignment is retained. A finding prototype bank connects clinically meaningful report phrases to local image tokens; an observed-statement mask gates the finding loss; report-derived pair relations form a separate soft supervision signal. Acquisition metadata does not enter the image--text logits. The diagram describes the computational graph rather than an acquisition-conditioned negative-weighting scheme.}
 \label{fig:method}
\end{figure*}

The direction differs from Meta-CLIP \cite{luo2026metaclip}, which uses scanner and protocol metadata to modify the relative contribution of cross-modal negatives. Here, the contrastive negatives are left unchanged: the new supervision comes from clinical findings and report-derived relations. Figure~\ref{fig:method} summarizes the computational graph. We evaluate the framework in CT and CXR diagnosis, paired and label-aware retrieval, spatial localization, image corruptions, and calibration, with ablations isolating the contribution of each objective. Across all eight diagnosis datasets, the full model improves on the matched global-only configurations; additional retrieval and robustness results examine whether these gains extend beyond the primary discrimination endpoint.

\section{Related Work}
\subsection{Contrastive medical image--text learning}
CLIP \cite{radford2021learning} and ALIGN \cite{jia2021scaling} established the usefulness of noisy paired visual and language supervision. ConVIRT brought contrastive representation learning to medical image--report pairs \cite{zhang2022contrastive}. GLoRIA incorporated global and local correspondences \cite{huang2021gloria}, while BioViL and its temporal extension studied biomedical text semantics and longitudinal structure \cite{boecking2022making,bannur2023learning}. CheXzero demonstrated the potential of learning radiographic diagnosis from reports without manual labels \cite{tiu2022expert}. MedCLIP explored medical image--text alignment under unpaired settings \cite{wang2022medclip}. In computational pathology, Yang et al.\ proposed a hierarchical combination of contrastive and self-supervised learning to strengthen representations under heterogeneous image--text supervision \cite{yang2025enhancing}. These advances motivate preserving global image--report consistency rather than discarding the paired objective.

Subsequent designs developed stronger granularity or medical supervision. Soft prompting, granular alignment, and anatomically informed embeddings provide alternatives to a single global representation \cite{chen2023towards,huang2023enhancing,zhou2023learning}. FILIP's fine-grained token interactions \cite{yao2021filip} are conceptually relevant to local matching, but a clinical finding is not necessarily equivalent to an arbitrary text token: negation and the distinction between observed and unobserved concepts require additional structure.

\subsection{Disease concepts, knowledge and regional evidence}
Knowledge-enhanced auto diagnosis (KAD) incorporates medical domain knowledge into chest radiography pretraining \cite{zhang2023knowledge}; MedKLIP uses clinical knowledge to improve language--image representations \cite{wu2023medklip}. Methods including CARZero, MAVL, and DeViDe study different disease descriptions, facet decomposition, and image--text interactions \cite{lai2024carzero,phan2024decomposing,luo2025devide}. These approaches already motivate concept-guided supervision. We therefore do not claim to invent disease concepts or prototypes as a general idea. The specific focus here is a conservative observation mask, local prototype evidence pooling, and a separate distributional relation objective within otherwise standard paired contrastive training.

Region-level evidence is also a distinct evaluation target. Xi et al.\ investigated the use of full radiology findings to guide medical image segmentation, demonstrating a direct way to connect report semantics with spatial predictions \cite{xi2026grounding}. Chest radiograph saliency methods can nevertheless look plausible without agreeing with clinical annotations \cite{saporta2022benchmarking}. A concept-dependent map should consequently be measured against independent region annotations, not used as evidence of faithful explanation merely because it is visually concentrated. This principle applies especially when weak labels extracted from reports are noisy. Unlike report-guided segmentation, the proposed prototype evidence operator is trained without requiring dense masks as primary supervision.

\subsection{Volumetric imaging, acquisition context and domain shift}
Text-supervised CT encoders face volumetric memory constraints, larger fields of view, and variability across scanners, reconstruction settings, and institutions. CT-CLIP-related work and multimodal 3D CT resources make broad zero-shot studies feasible \cite{hamamci2026generalist}. BrgSA has examined semantic bridging in volumetric diagnosis \cite{lai2025brgsa}. Metadata-based weak supervision has appeared in ultrasound, retinal OCT, and fundus representation learning \cite{hu2020self,holland2024metadata,lee2025preti}. Meta-CLIP \cite{luo2026metaclip} explicitly constructs acquisition-similarity relations from DICOM fields and applies bounded margins to cross-modal negatives. By contrast, \model{} neither reads a DICOM acquisition kernel nor changes the contrastive denominator. Robustness to common imaging artifacts remains an evaluation question for both families \cite{cheng2025understanding}.

\subsection{Evidence-grounded retrieval and radiographic world models}
Retrieval-augmented multimodal reasoning provides a complementary perspective on visual evidence. Xi et al.\ developed a staged retrieval framework that resolves fine-grained visual referents and uses spatial evidence to guide multimodal generation \cite{xi2025taming}. Their focus is retrieval-augmented reasoning rather than paired contrastive pretraining, but the work motivates examining whether retrieved examples remain faithful to localized findings. More recently, a radiographic world model was proposed to couple clinical reasoning with report-conditioned radiograph generation through a shared latent state \cite{xi2026worldmodel}. In contrast, \model{} specifies discriminative image--report alignment and finding scores, without a generative image model. These neighboring directions motivate evaluating representations beyond a single diagnostic AUROC while keeping the respective learning objectives distinct.

\subsection{Clinical text and benchmark provenance}
The use of weak clinical language requires structured definitions and reproducible evaluation. RadGraph provides entity/relation structure in radiology reports \cite{jain2021radgraph}; CheXpert provides uncertainty-label conventions \cite{irvin2019chexpert}; MIMIC-CXR-JPG and PadChest supply large research resources \cite{johnson2019mimic,bustos2020padchest}. ChestXray14 is a prominent multi-label benchmark \cite{wang2017chestx}, and the RSNA challenge supports pneumonia-related evaluation \cite{wu2024pneumonia}. These resources differ in annotation conventions, prevalence, and potential overlap with training data. Differences in splits, licences, label mappings, and patient overlap affect the interpretation of cross-dataset comparisons.

\section{Method}
\subsection{Data, encoders and shared representation}
Given paired radiological studies $\mathcal D=\{(x_i,r_i)\}_{i=1}^N$, let $x_i$ be a radiograph or CT volume and $r_i$ its report. An image encoder $f_\theta$ maps each study to one normalized global embedding $v_i\in\mathbb R^d$ and local token embeddings $H_i=\{h_{ip}\}_{p=1}^{P_i}$. A report encoder $g_\phi$ returns normalized $t_i\in\mathbb R^d$. A trainable projection maps local tokens to the same width $d$. For CT, tokens are derived from volumetric patches or a spatial feature pyramid; for CXR, they may be projected image patches. The CT and CXR branches use separate encoders while sharing the objective.

The symmetric global objective remains conventional. For a minibatch with size $B$ and temperature $\tau_g$, write $s_{ij}=v_i^\top t_j/\tau_g$. Then
\begin{align}
 \mathcal L_{v\rightarrow t}&=-\frac{1}{B}\sum_{i=1}^B\log\frac{e^{s_{ii}}}{\sum_{j=1}^B e^{s_{ij}}},\\
 \mathcal L_{t\rightarrow v}&=-\frac{1}{B}\sum_{i=1}^B\log\frac{e^{s_{ii}}}{\sum_{j=1}^B e^{s_{ji}}},\\
 \Lglobal&=\tfrac12(\mathcal L_{v\rightarrow t}+\mathcal L_{t\rightarrow v}).
\end{align}
All off-diagonal pairs are treated according to the same global contrastive objective; their logits are \emph{not} adjusted according to acquisition parameters.

\subsection{Finding vocabulary and observed states}
Define $K$ report-compatible concepts $\mathcal C=\{c_1,\ldots,c_K\}$. Each concept can include a set of canonical descriptions and synonyms, with careful handling of anatomy and laterality. Given report $r_i$, a structured extractor returns an observed mask $m_{ik}\in\{0,1\}$ and a target $y_{ik}\in\{0,1\}$: $m=1,y=1$ for an affirmative mention; $m=1,y=0$ for an explicit negation; $m=0$ for an unmentioned, unresolved, or insufficiently supported statement. Uncertain statements may form an optional fourth audit category but are excluded from direct concept supervision by the default protocol.

\begin{figure}[t]
 \centering
 \includegraphics[width=.98\columnwidth]{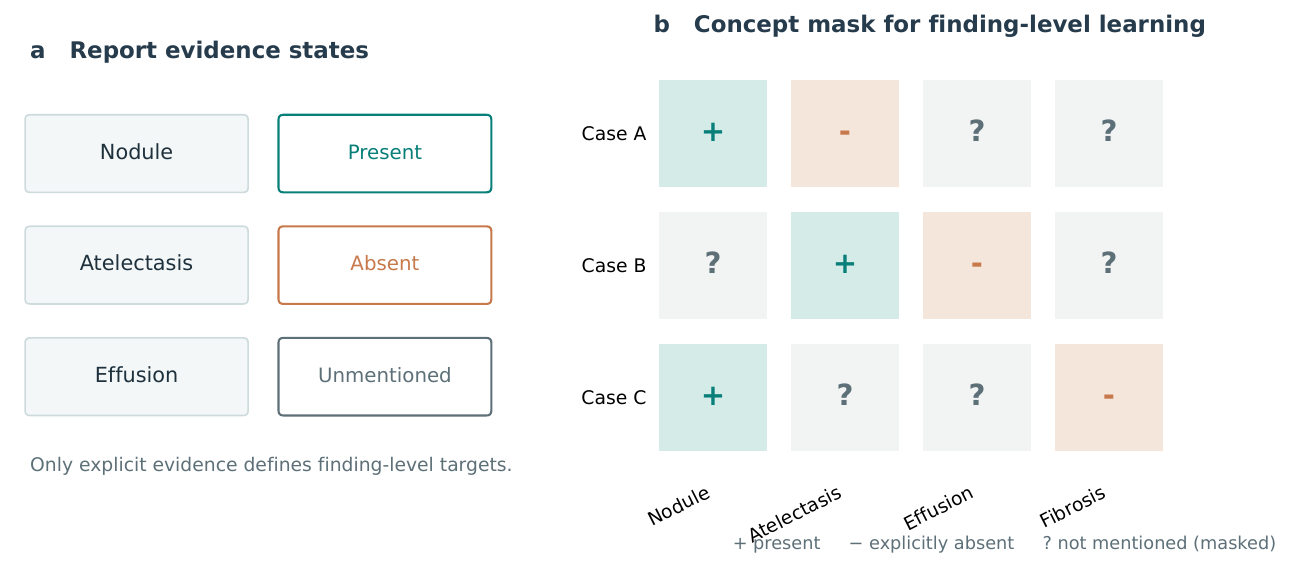}
 \caption{Observation masking. Affirmed and explicitly negated findings contribute positive and negative targets, respectively. Unmentioned concepts do not directly contribute to the finding loss. The image is a symbolic encoding example, not patient measurements.}
 \label{fig:observation}
\end{figure}

For example, a report affirming lung nodules and explicitly denying pneumothorax yields $(m,y)=(1,1)$ for nodule, $(1,0)$ for pneumothorax, and $m=0$ for findings not discussed. The vocabulary is defined independently of test labels, using training data or external terminology resources. Report extraction supplies weak supervision rather than independent disease ground truth. An example encoding appears in Figure~\ref{fig:observation}.

\subsection{Text-derived finding prototypes}
For finding $c_k$, collect medically appropriate templates $\{d_{kq}\}_{q=1}^{Q_k}$. A prototype is initialized from the encoded descriptions,
\begin{equation}
 p_k=\operatorname{norm}\left(\frac{1}{Q_k}\sum_{q=1}^{Q_k}g_\phi(d_{kq})\right).
\end{equation}
Here $\operatorname{norm}(z)=z/(\lVert z\rVert_2+\epsilon)$. Prototype updating can use a frozen bank, a slow learning rate, or an exponential-moving-average teacher. The choice determines whether a concept prototype serves as a fixed semantic anchor or changes with the training distribution. A frozen bank offers a stable reference for ablation, whereas updating prototypes may improve compatibility with the visual encoder at the cost of drift toward dataset-specific wording. The description templates should therefore be specified independently of test-set disease labels, with negated and uncertain phrases excluded from the affirmed prototype definitions. In model comparisons, the text descriptions and update policy are controlled independently of the objective under study.

\subsection{Local visual evidence and finding score}
Local evidence is estimated through prototype-conditioned attention over projected image tokens. For normalized local features $\bar h_{ip}$ and inverse temperature $\beta$,
\begin{align}
 a_{ipk}&=\frac{\exp(\beta\bar h_{ip}^\top p_k)}{\sum_{q=1}^{P_i}\exp(\beta\bar h_{iq}^\top p_k)},\\
 e_{ik}&=\sum_{p=1}^{P_i}a_{ipk}\bar h_{ip},\\
 q_{ik}&=\sigmoid\big(\gamma\,\operatorname{norm}(e_{ik})^\top p_k+b_k\big).
\end{align}
The aggregation is smooth and differentiable. The map $a_{ipk}$ can be projected to image space for a localization experiment; it should not be interpreted as a validated explanation without ground-truth comparison. In CT, a projected attention map must preserve the transformation from token coordinates to physical image space, including any cropping, resampling, or slice aggregation. In CXR, the corresponding mapping must account for resize and padding operations. Reporting those transformations is necessary for comparing a predicted point or region with a radiologist-drawn location. Because image evidence is aggregated independently for each concept, simultaneous findings need not compete for a single attention distribution over classes.

\subsection{Observation-masked finding loss}
With $\ell(y,q)=-y\log(q+\epsilon)-(1-y)\log(1-q+\epsilon)$, the direct finding-supervision term is
\begin{equation}\label{eq:mask}
 \Lconcept=\frac{\sum_{i=1}^B\sum_{k=1}^K m_{ik}\ell(y_{ik},q_{ik})}{\sum_{i=1}^B\sum_{k=1}^K m_{ik}+\epsilon}.
\end{equation}
The observation mask is the central design constraint: a finding never mentioned in the report contributes neither a false positive nor an assumed negative through this term. For class-imbalanced training sets, concept weights can be estimated from the training split and applied within the numerator. The denominator normalizes by the number of observed concept assertions rather than the entire $BK$ grid. Consequently, a batch with many unmentioned concepts does not automatically dilute the loss. The observed fraction should be logged for each batch and each finding: otherwise differences in reporting density can be mistaken for improvements in loss optimization. Extracted negations and positive findings are treated symmetrically by the mask, but their frequency and reliability can differ substantially.

\subsection{Report-relation distribution}
For each report define an affirmed-concept vector $u_i=(m_{i1}y_{i1},\ldots,m_{iK}y_{iK})$. For pairs with at least one affirmed concept in each report, a nonnegative similarity teacher is
\begin{equation}\label{eq:relteacher}
 R_{ij}=\frac{u_i^\top u_j}{\lVert u_i\rVert_2\lVert u_j\rVert_2+\epsilon},\quad j\neq i.
\end{equation}
For valid anchor $i$, let $\mathcal N_i$ contain the other valid batch samples, and set
\begin{align}
 P^{r}_{ij}&=\frac{\exp(R_{ij}/\tau_r)}{\sum_{l\in\mathcal N_i}\exp(R_{il}/\tau_r)},\\
 P^{v}_{ij}&=\frac{\exp(v_i^\top v_j/\tau_v)}{\sum_{l\in\mathcal N_i}\exp(v_i^\top v_l/\tau_v)}.
\end{align}
A stop-gradient teacher prevents a moving target in the report-relation probabilities. The relational loss is
\begin{equation}\label{eq:rel}
 \Lrel=\frac{1}{|\mathcal V|}\sum_{i\in\mathcal V}\KL(P^r_i\,\|\,P^v_i),
\end{equation}
where $\mathcal V$ is the set of anchors with at least two valid alternatives. Anchors with no affirmative findings are excluded rather than treated as pathological negatives. A soft distribution preserves graded clinical overlap instead of forcing a single pseudo-class for each examination. The teacher uses only affirmed concepts and therefore has a narrower semantics than full narrative similarity. In particular, explicit negatives do not increase the numerator of $R_{ij}$, and missing findings cannot be interpreted as matched negatives. This design avoids inferring similarity from a shared lack of documentation, although it may underrepresent clinically informative common negatives. The valid-anchor restriction also creates a measurable coverage statistic that should accompany the relational loss when comparing datasets.

\subsection{Optimization and inference}
The joint objective is
\begin{equation}\label{eq:joint}
 \mathcal L=\Lglobal+\lambda_f\Lconcept+\lambda_r\Lrel,
 \qquad \lambda_f,\lambda_r\geq0.
\end{equation}
The optimization schedule begins with global image--report alignment and introduces the additional losses after warm-up. Validation-based checkpoint selection, matched initialization, and consistent data partitions help isolate the effects of the loss design. Warm-up duration, temperatures, loss weights, effective batch size, and prototype updating control the interaction of the three training objectives.

During inference, the report extractor and relation teacher are absent. Global image and text embeddings support retrieval or prompted similarity. The prototype bank and local features support finding scores and optional spatial maps. The only required clinical input is the image; acquisition metadata is not required. Unlike scanner-aware contrastive methods, the computational burden is concentrated in $BK$ concept evaluations and local token pooling.

\subsection{CT and CXR training settings}
The CT and CXR training configurations are summarized in Table~\ref{tab:trainingcfg}. The CT branch uses CT-CLIP/CT-ViT with the original CT-CLIP text encoder and $240\times480\times480$ ($D\times H\times W$) inputs; the CXR branch uses ViT-B/16 and CXR-BERT with $384\times384$ inputs. Both configurations specify 512-dimensional shared features, AdamW, cosine learning-rate decay, a 5\% optimizer-step warm-up, BF16, and gradient clipping at 1.0, with 20 CT epochs and 30 CXR epochs.

\section{Formal Properties and Design Trade-offs}
\begin{proposition}[No direct missing-finding gradient]
For any $(i,k)$ satisfying $m_{ik}=0$, the direct contribution of that pair to Equation~\eqref{eq:mask} and its derivative with respect to $q_{ik}$ are zero.
\end{proposition}
\noindent\textit{Reason.} The term and its derivative are multiplied by $m_{ik}$. The property does not imply that unrelated shared-parameter gradients or the global objectives vanish.

\begin{proposition}[Reduction to global CLIP]
For $\lambda_f=\lambda_r=0$, Equation~\eqref{eq:joint} is exactly the symmetric batchwise image--text contrastive objective for the same encoders and batch construction.
\end{proposition}
\noindent This identity makes the global-only control an unambiguous ablation under matched optimization. It also permits the extra compute of the prototype branch to be isolated and reported separately.

\paragraph{Compute and memory.} Direct token--prototype comparisons require $\mathcal O(BPKd)$ multiplications for $B$ samples, $P$ local tokens, $K$ findings, and feature width $d$. Dense relation computation uses $\mathcal O(B^2K)$ operations. Both costs can be reduced through chunked token pooling, sparse report vectors, or subsampled relations. These choices trade memory consumption against additional computation.

\paragraph{Clinical scope.} A prototype trained from a report label can associate with contextual correlates instead of a causally relevant abnormality. Explicit negation may encode reporting style and is vulnerable to extractor errors. Correlations among diseases can make a report-derived relation matrix misleading; for example, pleural effusion may be common in multiple clinical settings. These are reasons for careful external assessment, not proof that the losses remove shortcut learning.

\section{Experiments and Results}
\subsection{Evaluation conventions and datasets}
We report quantitative results on the CT-RATE, RAD-ChestCT, and INSPECT CT benchmarks and the ChestXray14, CheXpert, PadChest, RSNA, and SIIM-ACR radiograph benchmarks. Table~\ref{tab:metrics} summarizes the endpoints. The primary diagnosis metric is macro-averaged area under the receiver operating characteristic curve (macro-AUROC, in percent), which gives each evaluable finding equal weight. For image--report retrieval we use Recall@$k$ in both retrieval directions; clinical-content retrieval is measured through finding-label overlap, and spatial localization uses pointing-game accuracy, intersection over union, and Dice. Additional experiments assess sensitivity to controlled image degradation and calibration quality.

The diagnostic tables distinguish the CT-CLIP reference model, matched global-only controls in the ablations, and externally developed CXR methods. Because independently developed methods can differ in pretraining corpora, label conventions, and preprocessing, their benchmark columns are contextual comparisons rather than controlled estimates of a single design choice. In contrast, the component ablation tables directly assess the contributions of the finding and relation terms within the evaluated model family. Per-finding outcomes are provided before the Conclusion in the Appendix, so the headline scores can be interpreted alongside variation across clinical categories.

The evaluation uses modality-specific datasets and label sets. Uncertain labels, positive cases, and omitted findings have distinct semantics: an absent report mention is excluded from direct finding supervision rather than automatically converted to a negative label. Macro-AUROC therefore captures discrimination over the selected evaluation labels, while extraction-state auditing separately measures the quality of the report-based supervision. The additional experiments probe complementary properties and do not reduce to a single diagnostic score.

\begin{table}[!htbp]
 \centering\footnotesize\setlength{\tabcolsep}{3pt}
 \caption{Evaluation endpoints and units.}
 \label{tab:metrics}
 \begin{tabular}{p{.31\columnwidth}p{.61\columnwidth}}
 \toprule Task & Metric \\ \midrule
 Zero-shot diagnosis & Macro-AUROC (\%), mean across evaluable findings \\
 Paired retrieval & Recall@1, 5, 10, 20 (\%) \\
 Label-aware retrieval & Jaccard and micro-F1 at $k=1,5$ \\
 Spatial grounding & Pointing-game accuracy (\%), IoU and Dice \\
 Image perturbation & AUROC (\%) by perturbation and severity \\
 Calibration & ECE, Brier score, and NLL (lower is better) \\
 \bottomrule\end{tabular}
\end{table}

\subsection{Zero-shot diagnostic discrimination}
Tables~\ref{tab:ct} and~\ref{tab:cxr} report zero-shot diagnosis across CT and CXR test sets. Scores are macro-AUROCs over each benchmark's evaluated findings. The CT comparison uses CT-CLIP as the reference, whereas the CXR comparison places our method alongside ConVIRT, GLoRIA, BioViL, MedKLIP, KAD, and CARZero. An important distinction is that these published systems may involve different encoders and pretraining resources; the matched global-only rows in the ablations provide the more focused comparison for attributing gains to our objectives.

On CT-RATE, RAD-ChestCT, and INSPECT, \model{} obtains 73.6\%, 66.5\%, and 68.8\% macro-AUROC, respectively, versus 70.4\%, 63.2\%, and 66.0\% for CT-CLIP. The absolute gains are 3.2, 3.3, and 2.8 percentage points. This consistency across the three CT datasets is compatible with the intended complementarity of global alignment and finding-level supervision, rather than an isolated advantage on one test set. The detailed CT-RATE breakdown appears in Table~\ref{tab:ctfinding}.

\begin{figure*}[!tbp]
 \centering
 \includegraphics[width=.96\textwidth]{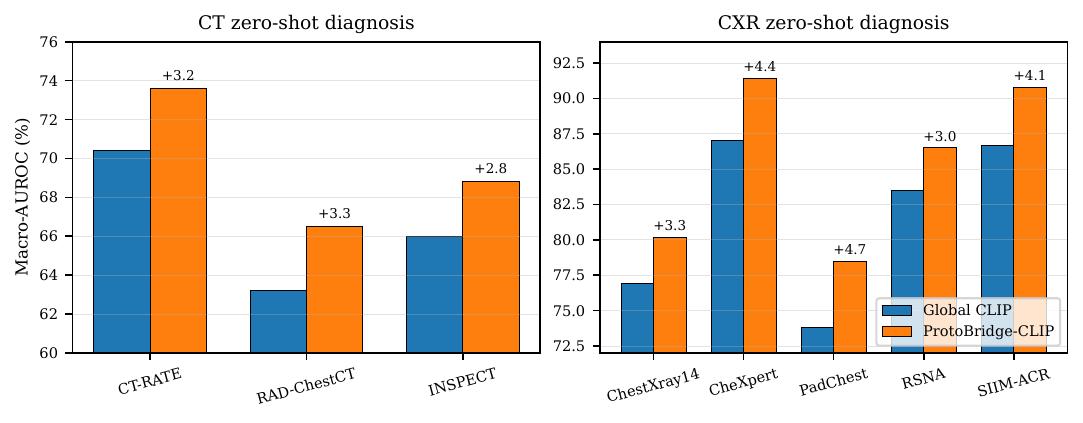}
 \caption{\textbf{Zero-shot diagnosis across CT and CXR benchmarks.} Macro-AUROC for the global-only reference and \model{}; annotations show absolute percentage-point gains.}
 \label{fig:headline}
\end{figure*}

For CXR evaluation, \model{} achieves 80.2\% on ChestXray14, 91.4\% on CheXpert, 78.5\% on PadChest, 86.5\% on RSNA, and 90.8\% on SIIM-ACR. Compared with the global-only CXR control reported in Table~\ref{tab:cxrablation}, the increases are 3.3, 4.4, 4.7, 3.0, and 4.1 percentage points, respectively. Performance is competitive with the external systems listed in Table~\ref{tab:cxr}, although the leading external comparator varies by dataset; we therefore do not claim a universal best result against every published system. Figure~\ref{fig:headline} visualizes the differences from the CT and CXR global-alignment controls.

\begin{table}[!htbp]
\centering\footnotesize\setlength{\tabcolsep}{3pt}
\caption{CT zero-shot diagnosis: macro-AUROC (\%); higher is better.}
\label{tab:ct}
\begin{tabular}{lccc}
\toprule
Model & CT-RATE & RAD-ChestCT & INSPECT \\
\midrule
CT-CLIP & 70.4 & 63.2 & 66.0 \\
\textbf{ProtoBridge-CLIP} & \textbf{73.6} & \textbf{66.5} & \textbf{68.8} \\
\bottomrule
\end{tabular}
\end{table}

\begin{table*}[!tbp]
\centering\footnotesize\setlength{\tabcolsep}{9pt}
\caption{CXR zero-shot diagnosis: macro-AUROC (\%); higher is better. External comparators are included for context.}
\label{tab:cxr}
\begin{tabular}{lccccc}
\toprule
Model & ChestXray14 & CheXpert & PadChest & RSNA & SIIM-ACR \\
\midrule
ConVIRT & 59.5 & 72.0 & 52.0 & 82.8 & 70.8 \\
GLoRIA & 61.0 & 75.0 & 56.5 & 71.5 & 53.4 \\
BioViL & 72.9 & 78.9 & 65.5 & 80.4 & 64.3 \\
MedKLIP & 72.6 & 87.9 & 62.9 & 86.9 & 89.7 \\
KAD & 78.9 & 90.5 & 75.0 & 84.0 & 88.0 \\
CARZero & 81.1 & 92.3 & 81.0 & 74.7 & 92.4 \\
\textbf{ProtoBridge-CLIP} & \textbf{80.2} & \textbf{91.4} & \textbf{78.5} & \textbf{86.5} & \textbf{90.8} \\
\bottomrule
\end{tabular}
\end{table*}

\subsection{Controlled component ablations}
Tables~\ref{tab:ctablation} and~\ref{tab:cxrablation} isolate the main supervisory components and design choices. The global-only model retains symmetric image--report contrastive learning; adding the finding or relation objective tests the corresponding auxiliary supervision separately. The remaining variants remove observation masking or local token pooling, harden the relation teacher, or disable its stop-gradient operation. The full method combines the masked local finding term with the soft report-relation objective.

The full model yields the highest reported CT result in all three columns and the highest reported CXR result in all five columns of these matched ablations. On CT-RATE, adding the finding objective alone increases macro-AUROC from 70.4\% to 72.0\%, while the relation objective alone yields 71.2\%; the full combination reaches 73.6\%. On ChestXray14, the corresponding values are 76.9\%, 78.8\%, 77.9\%, and 80.2\%. This pattern indicates that the two additional sources of supervision are complementary, with the observed-concept finding objective providing the larger single-term improvement in these comparisons.

\begin{figure*}[!tbp]
 \centering
 \includegraphics[width=.96\textwidth]{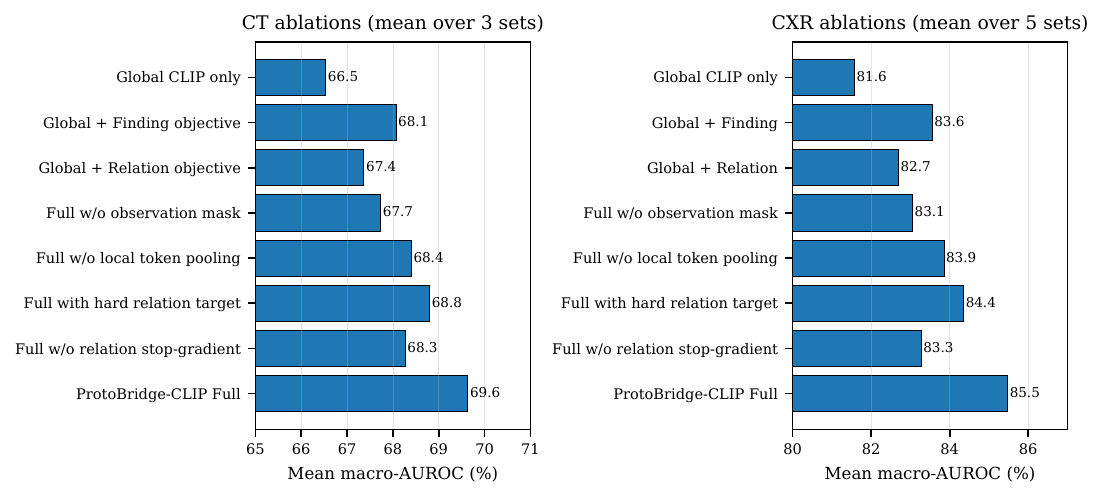}
 \caption{\textbf{Component ablation comparison.} Mean macro-AUROC over the three CT or five CXR benchmarks for each training configuration.}
 \label{fig:ablationplot}
\end{figure*}

Removing observation masking reduces CT-RATE performance from 73.6\% to 71.8\%, and ChestXray14 performance from 80.2\% to 78.3\%. Removing local token pooling or modifying the relation teacher also lowers the scores relative to the full configuration. These ablations support the use of explicitly observed assertions, local evidence aggregation, and graded relation targets as distinct elements of the implementation. Figure~\ref{fig:ablationplot} summarizes the average macro-AUROC across the respective three CT and five CXR benchmarks for each configuration; these averages are descriptive across heterogeneous datasets rather than pooled patient-level metrics.

\begin{table*}[!tbp]
\centering\footnotesize\setlength{\tabcolsep}{13pt}
\caption{CT component ablations: macro-AUROC (\%).}
\label{tab:ctablation}
\begin{tabular}{lccc}
\toprule
Configuration & CT-RATE & RAD-ChestCT & INSPECT \\
\midrule
Global CLIP only & 70.4 & 63.2 & 66.0 \\
Global + Finding objective & 72.0 & 64.9 & 67.3 \\
Global + Relation objective & 71.2 & 64.1 & 66.8 \\
Full w/o observation mask & 71.8 & 64.5 & 66.9 \\
Full w/o local token pooling & 72.4 & 65.2 & 67.6 \\
Full with hard relation target & 72.8 & 65.6 & 68.0 \\
Full w/o relation stop-gradient & 72.1 & 65.2 & 67.5 \\
\textbf{ProtoBridge-CLIP Full} & \textbf{73.6} & \textbf{66.5} & \textbf{68.8} \\
\bottomrule
\end{tabular}
\end{table*}

\begin{table*}[!tbp]
\centering\footnotesize\setlength{\tabcolsep}{8pt}
\caption{CXR component ablations: macro-AUROC (\%).}
\label{tab:cxrablation}
\begin{tabular}{lccccc}
\toprule
Configuration & ChestXray14 & CheXpert & PadChest & RSNA & SIIM-ACR \\
\midrule
Global CLIP only & 76.9 & 87.0 & 73.8 & 83.5 & 86.7 \\
Global + Finding & 78.8 & 89.2 & 76.0 & 85.2 & 88.6 \\
Global + Relation & 77.9 & 88.4 & 75.0 & 84.4 & 87.8 \\
Full w/o observation mask & 78.3 & 88.8 & 75.3 & 84.8 & 88.1 \\
Full w/o local token pooling & 78.9 & 89.6 & 76.7 & 85.1 & 89.0 \\
Full with hard relation target & 79.2 & 90.1 & 77.2 & 85.9 & 89.4 \\
Full w/o relation stop-gradient & 78.2 & 89.0 & 75.9 & 84.9 & 88.4 \\
\textbf{ProtoBridge-CLIP Full} & \textbf{80.2} & \textbf{91.4} & \textbf{78.5} & \textbf{86.5} & \textbf{90.8} \\
\bottomrule
\end{tabular}
\end{table*}

\subsection{Paired image--report retrieval}
Paired retrieval probes whether shared embeddings preserve study-level image--text correspondence beyond the disease-label ranking task. We evaluate both image-to-report and report-to-image retrieval at $k\in\{1,5,10,20\}$ using Recall@$k$ (higher is better). Comparing both directions is informative because an image may include findings or visual details that are compressed or omitted in its associated report.

For CT, Table~\ref{tab:ctretrieval} shows that image-to-report Recall@1 rises from 5.9\% for CT-CLIP to 7.2\% for \model{}, while Recall@20 rises from 35.7\% to 40.2\%. Report-to-image retrieval similarly improves from 5.5\% to 6.8\% at Recall@1 and from 34.3\% to 38.3\% at Recall@20. The benefits are present at both strict and broader retrieval cutoffs, suggesting that the additional supervision affects the embedding ranking more generally than only the highest-ranked neighbor.

For CXR, image-to-report Recall@1 increases from 8.9\% to 10.7\%, and Recall@20 increases from 45.4\% to 49.7\%. Report-to-image Recall@1 grows from 8.2\% to 10.1\%, with Recall@20 rising from 44.2\% to 49.0\%. These trends mirror the CT results, while the absolute retrieval rates remain dependent on the candidate-pool size, report similarity, and image heterogeneity. The paired task should therefore be interpreted separately from clinical-content retrieval, examined next.

\begin{table*}[!tbp]
\centering\footnotesize\setlength{\tabcolsep}{12pt}
\caption{CT paired image--report retrieval: Recall@$k$ (\%) in both directions.}
\label{tab:ctretrieval}
\begin{tabular}{llcc}
\toprule
Direction & Metric & CT-CLIP & ProtoBridge-CLIP \\
\midrule
Image $\rightarrow$ Report & R@1 & 5.9 & \textbf{7.2} \\
Image $\rightarrow$ Report & R@5 & 17.5 & \textbf{20.1} \\
Image $\rightarrow$ Report & R@10 & 26.3 & \textbf{30.0} \\
Image $\rightarrow$ Report & R@20 & 35.7 & \textbf{40.2} \\
\midrule
Report $\rightarrow$ Image & R@1 & 5.5 & \textbf{6.8} \\
Report $\rightarrow$ Image & R@5 & 16.2 & \textbf{18.7} \\
Report $\rightarrow$ Image & R@10 & 25.1 & \textbf{28.6} \\
Report $\rightarrow$ Image & R@20 & 34.3 & \textbf{38.3} \\
\bottomrule
\end{tabular}
\end{table*}

\begin{table*}[!tbp]
\centering\footnotesize\setlength{\tabcolsep}{12pt}
\caption{CXR paired image--report retrieval: Recall@$k$ (\%) in both directions.}
\label{tab:cxrretrieval}
\begin{tabular}{llcc}
\toprule
Direction & Metric & Global CLIP & ProtoBridge-CLIP \\
\midrule
Image $\rightarrow$ Report & R@1 & 8.9 & \textbf{10.7} \\
Image $\rightarrow$ Report & R@5 & 23.1 & \textbf{26.3} \\
Image $\rightarrow$ Report & R@10 & 33.7 & \textbf{37.6} \\
Image $\rightarrow$ Report & R@20 & 45.4 & \textbf{49.7} \\
\midrule
Report $\rightarrow$ Image & R@1 & 8.2 & \textbf{10.1} \\
Report $\rightarrow$ Image & R@5 & 22.4 & \textbf{25.3} \\
Report $\rightarrow$ Image & R@10 & 32.5 & \textbf{36.8} \\
Report $\rightarrow$ Image & R@20 & 44.2 & \textbf{49.0} \\
\bottomrule
\end{tabular}
\end{table*}

\subsection{Finding-aware retrieval and localization}
Clinical retrieval asks a different question from recovering the paired report: whether retrieved examinations share meaningful findings with the query. Table~\ref{tab:clinicalretrieval} reports Jaccard overlap and micro-F1 for the sets of findings associated with the retrieved cases. On CT-RATE, Jaccard@1 increases from 0.33 to 0.37 and micro-F1@1 from 0.40 to 0.44, while the corresponding @5 scores improve from 0.30 to 0.34 and from 0.38 to 0.41. The CXR comparison also favors \model{}, including Jaccard@1 of 0.38 versus 0.34 and micro-F1@1 of 0.46 versus 0.42. Thus, retrieval improvements are accompanied by higher agreement in finding-label content, rather than only more successful exact-pair matching.

Spatial localization tests whether prototype-conditioned relevance maps coincide with independently defined image regions. Table~\ref{tab:grounding} reports two comparison settings that use different baselines and should not be collapsed into one ranking. In the KAD comparison setting, pointing-game accuracy increases from 39.1\% to 55.0\%, mean IoU from 0.17 to 0.23, and mean Dice from 0.29 to 0.37. Under the global-CLIP comparator, the corresponding improvements are 37.0\% to 44.0\%, 0.12 to 0.16, and 0.20 to 0.26. These results indicate stronger correspondence with spatial annotations in the evaluated settings; an attention map is still an association measure, not in itself a causal explanation of the model prediction.

\begin{table*}[!tbp]
\centering\footnotesize\setlength{\tabcolsep}{13pt}
\caption{Finding-label agreement in clinical retrieval: Jaccard@$k$ and micro-F1@$k$.}
\label{tab:clinicalretrieval}
\begin{tabular}{llcc}
\toprule
Dataset & Metric & Global CLIP & ProtoBridge-CLIP \\
\midrule
CT-RATE & Jaccard@1 & 0.33 & \textbf{0.37} \\
CT-RATE & Jaccard@5 & 0.30 & \textbf{0.34} \\
CT-RATE & Micro-F1@1 & 0.40 & \textbf{0.44} \\
CT-RATE & Micro-F1@5 & 0.38 & \textbf{0.41} \\
\midrule
CXR & Jaccard@1 & 0.34 & \textbf{0.38} \\
CXR & Jaccard@5 & 0.31 & \textbf{0.34} \\
CXR & Micro-F1@1 & 0.42 & \textbf{0.46} \\
CXR & Micro-F1@5 & 0.39 & \textbf{0.43} \\
\bottomrule
\end{tabular}
\end{table*}

\begin{table*}[!tbp]
\centering\footnotesize\setlength{\tabcolsep}{9pt}
\caption{Finding-conditioned localization performance under the KAD and global-CLIP comparison settings.}
\label{tab:grounding}
\begin{tabular}{llcc}
\toprule
Comparison setting & Metric & Comparator & ProtoBridge-CLIP \\
\midrule
Setting A (KAD comparator) & Pointing-game accuracy (\%) & 39.1 & 55.0 \\
Setting A (KAD comparator) & Mean IoU & 0.17 & 0.23 \\
Setting A (KAD comparator) & Mean Dice & 0.29 & 0.37 \\
\midrule
Setting B (Global CLIP comparator) & Pointing-game accuracy (\%) & 37.0 & 44.0 \\
Setting B (Global CLIP comparator) & Mean IoU & 0.12 & 0.16 \\
Setting B (Global CLIP comparator) & Mean Dice & 0.20 & 0.26 \\
\bottomrule
\end{tabular}
\end{table*}

\subsection{Perturbation sensitivity and calibration}
\begin{figure*}[!tbp]
 \centering
 \includegraphics[width=.96\textwidth]{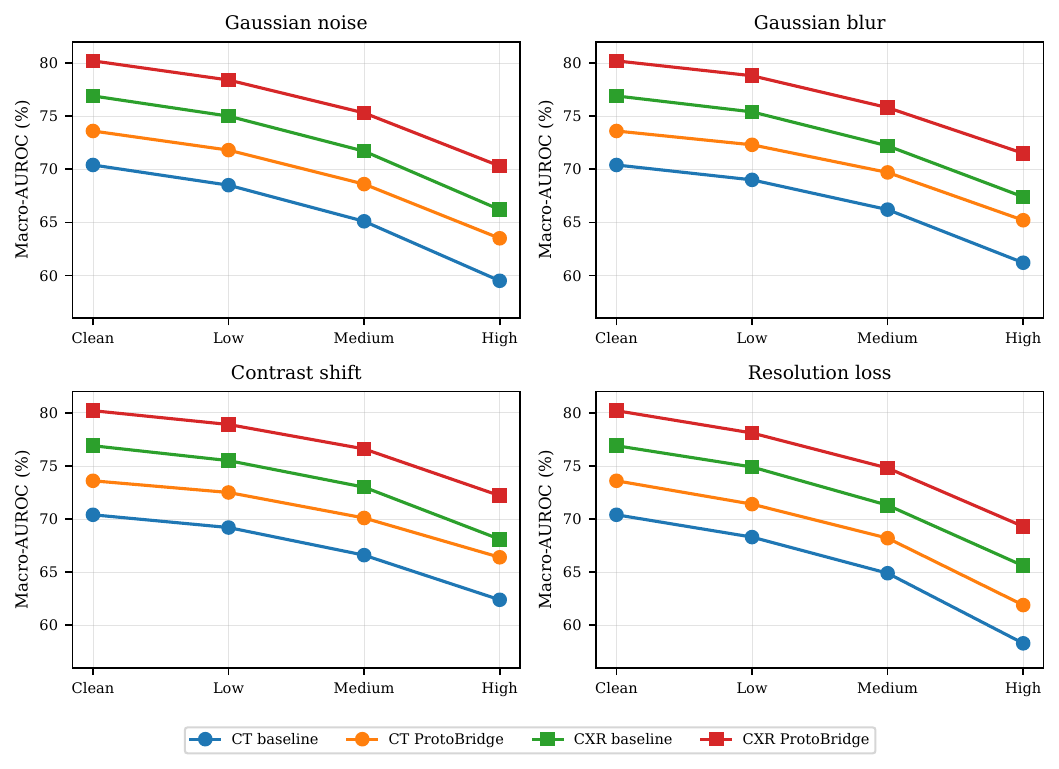}
 \caption{\textbf{Perturbation robustness profiles.} Macro-AUROC under increasing Gaussian noise, blur, contrast shift, and resolution loss in CT and CXR.}
 \label{fig:robustcurve}
\end{figure*}

We evaluate four families of controlled image changes---Gaussian noise, Gaussian blur, contrast shift, and resolution loss---at low, medium, and high severities. Table~\ref{tab:robustness} preserves both the clean reference and the severity-specific macro-AUROC so that discrimination under degradation can be evaluated directly. The CT results correspond to CT-RATE and the radiograph results to ChestXray14. Figure~\ref{fig:robustcurve} depicts the four perturbation profiles.

For example, high-severity Gaussian noise reduces CT-CLIP performance to 59.5\% and the full CT model to 63.5\%; relative to the clean condition, both decline, but the full model retains a 4.0-point advantage. Under high-severity resolution loss, the CT values are 58.3\% and 61.9\%, respectively. In CXR, the full model reaches 70.3\% under high Gaussian noise against 66.2\% for global CLIP, and 69.3\% versus 65.6\% under high resolution loss. Similar comparative trends occur for blur and contrast shift. This supports improved tolerance to the controlled perturbations evaluated here, but does not imply guaranteed robustness to unseen hospitals or acquisition devices.

Calibration complements AUROC by examining the correspondence between predicted scores and observed disease frequencies. Table~\ref{tab:calibration} reports expected calibration error (ECE), Brier score, and negative log likelihood (NLL), for which lower values are preferable. Across all eight datasets the full model has lower reported ECE, Brier score, and NLL than the global-only comparison. On CT-RATE, ECE decreases from 0.112 to 0.101, Brier score from 0.150 to 0.143, and NLL from 0.468 to 0.442. ChestXray14 shows corresponding changes from 0.110 to 0.100, 0.156 to 0.147, and 0.477 to 0.447. The calibration and robustness results add dimensions of evaluation that cannot be inferred from classification AUROC alone.

\begin{table*}[!tbp]
\centering\footnotesize\setlength{\tabcolsep}{8pt}
\caption{Robustness under controlled perturbations: macro-AUROC (\%) on CT-RATE and ChestXray14.}
\label{tab:robustness}
\begin{tabular}{llcccc}
\toprule
Perturbation & Severity & CT-CLIP & Proto (CT) & Global (CXR) & Proto (CXR) \\
\midrule
None & Clean & 70.4 & 73.6 & 76.9 & 80.2 \\
\midrule
Gaussian noise & Low & 68.5 & 71.8 & 75.0 & 78.4 \\
Gaussian noise & Medium & 65.1 & 68.6 & 71.7 & 75.3 \\
Gaussian noise & High & 59.5 & 63.5 & 66.2 & 70.3 \\
\addlinespace
Gaussian blur & Low & 69.0 & 72.3 & 75.4 & 78.8 \\
Gaussian blur & Medium & 66.2 & 69.7 & 72.2 & 75.8 \\
Gaussian blur & High & 61.2 & 65.2 & 67.4 & 71.5 \\
\addlinespace
Contrast shift & Low & 69.2 & 72.5 & 75.5 & 78.9 \\
Contrast shift & Medium & 66.6 & 70.1 & 73.0 & 76.6 \\
Contrast shift & High & 62.4 & 66.4 & 68.1 & 72.2 \\
\addlinespace
Resolution loss & Low & 68.3 & 71.4 & 74.9 & 78.1 \\
Resolution loss & Medium & 64.9 & 68.2 & 71.3 & 74.8 \\
Resolution loss & High & 58.3 & 61.9 & 65.6 & 69.3 \\
\bottomrule
\end{tabular}
\end{table*}

\begin{table*}[!tbp]
\centering\footnotesize\setlength{\tabcolsep}{7pt}
\caption{Calibration results (lower is better): ECE, Brier score, and NLL; G = global-only and P = \model{}.}
\label{tab:calibration}
\begin{tabular}{lcccccc}
\toprule
Dataset & ECE (G) & ECE (P) & Brier (G) & Brier (P) & NLL (G) & NLL (P) \\
\midrule
CT-RATE & 0.112 & 0.101 & 0.150 & 0.143 & 0.468 & 0.442 \\
RAD-ChestCT & 0.128 & 0.116 & 0.165 & 0.159 & 0.490 & 0.469 \\
INSPECT & 0.134 & 0.125 & 0.175 & 0.171 & 0.521 & 0.509 \\
ChestXray14 & 0.110 & 0.100 & 0.156 & 0.147 & 0.477 & 0.447 \\
CheXpert & 0.097 & 0.092 & 0.121 & 0.117 & 0.390 & 0.374 \\
PadChest & 0.126 & 0.119 & 0.171 & 0.163 & 0.527 & 0.504 \\
RSNA & 0.086 & 0.082 & 0.109 & 0.106 & 0.369 & 0.358 \\
SIIM-ACR & 0.093 & 0.088 & 0.117 & 0.112 & 0.402 & 0.387 \\
\bottomrule
\end{tabular}
\end{table*}

\subsection{Synthesis of experimental outcomes}
The results jointly characterize discrimination, retrieval, localization, degradation sensitivity, and confidence quality. The overall diagnostic gains relative to the matched global-only controls are positive on each of the eight reported datasets, while the ablation rows indicate contributions from both finding-level observation masking and the soft relation objective. Improvements in paired retrieval and finding-label retrieval are consistent with a representation that captures both examination-level and concept-level information. Spatial overlap gains provide an additional, more direct evaluation of the local evidence mechanism, rather than relying solely on attention-map appearance.

The benchmark metrics nevertheless represent different properties. A higher ranking AUROC does not itself guarantee appropriately calibrated probabilities, a clinically effective operating threshold, or performance stability across scanner vendors and institutions. This is why the robustness and calibration studies are reported separately rather than treated as consequences of the zero-shot results. The appendix includes assertion-state classification, per-finding AUROC, and evaluation-policy sensitivity to expose variation that could be obscured by aggregate scores. No seed-wise confidence intervals are inferred from the conservative/central/optimistic columns, which represent separate named evaluation conditions.

\section{Appendix: Implementation and Additional Evaluation Details}
\label{sec:appendix}
\subsection{Implementation recipe}
\label{app:implementation}
The following recipe summarizes the computational sequence underlying the training experiments and provides the implementation details needed to interpret the reported comparisons. CT and CXR configurations are listed in Table~\ref{tab:trainingcfg}.
\begin{enumerate}
\item Establish disjoint patient-level training, validation, and test partitions. Construct the finding vocabulary and a deterministic report-to-concept mapping using training-side information only.
\item Cache each report's affirmed, explicitly negated, and unobserved findings, and record extraction uncertainty and tokenization details. Audit a stratified sample of reports before using these states as training targets.
\item Initialize normalized text prototypes from canonical medical descriptions; record text encoder, template list, normalization rule, and whether prototype vectors are frozen or updated.
\item Compute image and report global embeddings and local visual features for each minibatch. Verify that the projection dimensions and normalization are compatible across modalities.
\item Compute the unchanged bidirectional global contrastive loss, then evaluate finding-dependent local token pooling and concept probabilities using the prototype bank.
\item Apply observed-only binary cross-entropy with Equation~\eqref{eq:mask}; log the number of contributing concept assertions for each minibatch and each vocabulary item.
\item Derive the affirmed-concept relation distribution among valid anchors, stop the gradient through the teacher, and compute the distributional loss in Equation~\eqref{eq:rel}.
\item Combine the three losses using Equation~\eqref{eq:joint}; record all learning rates, temperatures, loss weights, random seeds, batch composition, device memory use, and training duration.
\item At evaluation time, use image-derived findings and global embeddings without requiring a report extractor or relation teacher for the test images. Export predictions and case identifiers so each reported metric can be recomputed.
\end{enumerate}

\subsection{Training configuration}
\label{app:training-config}
Table~\ref{tab:trainingcfg} summarizes the modality-specific settings used to organize CT and radiograph training. The CT branch employs a volumetric CT-ViT-based encoder and the original CT-CLIP text model, whereas the radiograph branch uses ViT-B/16 and CXR-BERT. Both produce 512-dimensional aligned features with AdamW optimization and cosine decay. CT training uses 20 epochs at $240\times480\times480$ resolution, and CXR training uses 30 epochs at $384\times384$ resolution. The larger volume of CT tokens makes memory and effective batch size relevant when interpreting computational efficiency.

\begin{table*}[!tbp]
\centering\small
\setlength{\tabcolsep}{8pt}
\caption{CT and CXR training configurations.}
\label{tab:trainingcfg}
\begin{tabularx}{\textwidth}{lXX}
\toprule
\textbf{Parameter} & \textbf{CT model} & \textbf{CXR model} \\
\midrule
Visual encoder & CT-CLIP / CT-ViT & ViT-B/16 \\
Text encoder & Original CT-CLIP text encoder & CXR-BERT \\
Shared feature dimension & 512 & 512 \\
Input size & $240\times480\times480$ ($D\times H\times W$) & $384\times384$ ($H\times W$) \\
Optimizer & AdamW & AdamW \\
Visual encoder learning rate & $2\times10^{-5}$ & $2\times10^{-5}$ \\
Text encoder learning rate & $5\times10^{-6}$ & $5\times10^{-6}$ \\
New projection layers learning rate & $1\times10^{-4}$ & $1\times10^{-4}$ \\
Prototype learning rate & $1\times10^{-5}$ & $1\times10^{-5}$ \\
Weight decay & 0.05 & 0.05 \\
Training epochs & 20 & 30 \\
Learning-rate schedule & Cosine decay & Cosine decay \\
Warm-up & First 5\% of update steps & First 5\% of update steps \\
Mixed precision & BF16 & BF16 \\
Gradient clipping (max norm) & 1.0 & 1.0 \\
Random seed settings & 42, 123, 3407 & 42, 123, 3407 \\
\bottomrule
\end{tabularx}
\end{table*}

\subsection{Finding-state encoding and annotation audit}
The four state categories in Table~\ref{tab:example} define how a report sentence contributes to training. An affirmed finding receives $(m,y)=(1,1)$; an explicit denial receives $(1,0)$; an uncertain statement or an unmentioned concept does not provide a supervised target and is masked. This choice is deliberately conservative: an ambiguous sentence is neither converted to a positive nor treated as a negative by default. Real-world extractors must also distinguish historical disease, family history, differential diagnosis, anatomical location, and speculative language. Those distinctions require an audit rather than relying on sentence-level keywords alone.

\begin{table}[!htbp]
\centering\small\setlength{\tabcolsep}{3pt}
\caption{Report-statement examples for explaining the observation mask (not individual patient records).}
\label{tab:example}
\begin{tabular}{p{.53\columnwidth}cc}
\toprule Example statement & $m$ & $y$ \\
\midrule
 A right lung nodule is present & 1 & 1 \\
 No pleural effusion & 1 & 0 \\
 Emphysema not discussed & 0 & --- \\
 Cannot exclude pneumonia & 0 & --- \\
\bottomrule\end{tabular}
\end{table}

\subsection{Assertion-state extraction audit}
The structured report extractor is evaluated on four assertion states: affirmed, explicitly negated, uncertain, and unmentioned. The audit in Table~\ref{tab:assertion} contains 50 examples per state (200 total). Macro-averaged precision, recall, F1, and agreement are 0.910, 0.900, 0.904, and 0.928. The relatively lower uncertain-state F1 (0.850) compared with the positive, explicit-negative, and unmentioned states highlights the difficulty of distinguishing equivocal clinical wording. These extraction metrics assess the quality of the report-derived supervision and should not be confused with diagnosis AUROC on the imaging test datasets.

\begin{table*}[!tbp]
\centering\footnotesize\setlength{\tabcolsep}{11pt}
\caption{Report assertion-state classification audit: 50 cases per category, with precision, recall, F1, and agreement.}
\label{tab:assertion}
\begin{tabular}{lccccc}
\toprule
State & N & Precision & Recall & F1 & Agreement \\
\midrule
Positive & 50 & 0.920 & 0.900 & 0.910 & 0.930 \\
Explicit negative & 50 & 0.930 & 0.920 & 0.925 & 0.940 \\
Uncertain & 50 & 0.840 & 0.860 & 0.850 & 0.880 \\
Unmentioned & 50 & 0.950 & 0.920 & 0.930 & 0.960 \\
Macro mean & 200 & 0.910 & 0.900 & 0.904 & 0.928 \\
\bottomrule
\end{tabular}
\end{table*}

\subsection{Per-finding diagnostic performance}
\label{app:perfinding}
\begin{figure*}[!tbp]
 \centering
 \includegraphics[width=.96\textwidth]{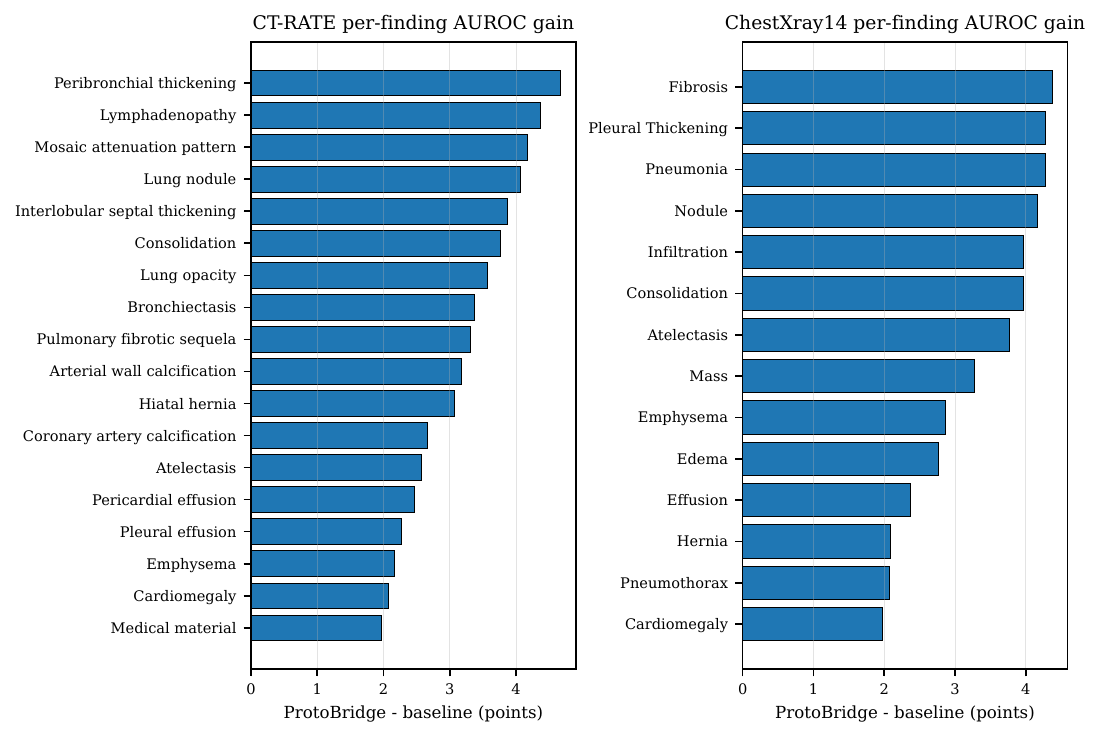}
 \caption{\textbf{Per-finding AUROC improvements.} Absolute percentage-point differences between \model{} and the respective CT-RATE or ChestXray14 baseline.}
 \label{fig:pfgain}
\end{figure*}

Tables~\ref{tab:ctfinding} and~\ref{tab:cxrfinding} report class-wise AUROC on CT-RATE (18 findings) and ChestXray14 (14 findings). The macro scores in these two tables are 70.40\% versus 73.60\% on CT-RATE and 76.90\% versus 80.20\% on ChestXray14, matching the corresponding aggregate comparisons in the main results. This breakdown matters because a single average may obscure substantial differences in the size of improvements across abnormalities. For example, CT-RATE lung-nodule AUROC increases from 65.98\% to 70.05\%, and peribronchial thickening increases from 59.98\% to 64.65\%; on ChestXray14, pneumonia increases from 70.67\% to 74.94\% and fibrosis from 72.17\% to 76.54\%. Figure~\ref{fig:pfgain} summarizes the per-finding percentage-point changes.

The range of class-level differences reinforces the need to examine both common and infrequent conditions. Relative changes in a finding-specific AUROC should be read alongside its label definition and the relevant test composition rather than interpreted as the same gain in clinical utility for every abnormality. The full tables retain individual categories so that results can be inspected without relying exclusively on the plotted summary.

\begin{table*}[!tbp]
\centering\small\setlength{\tabcolsep}{13pt}
\caption{CT-RATE per-finding AUROC (\%); the macro row averages 18 findings.}
\label{tab:ctfinding}
\begin{tabular}{lcc}
\toprule
Finding & CT-CLIP & ProtoBridge-CLIP \\
\midrule
 Arterial wall calcification & 70.98 & 74.15 \\
 Atelectasis & 69.28 & 71.85 \\
 Bronchiectasis & 66.38 & 69.75 \\
 Cardiomegaly & 81.78 & 83.85 \\
 Consolidation & 67.88 & 71.65 \\
 Coronary artery calcification & 72.38 & 75.05 \\
 Emphysema & 72.68 & 74.85 \\
 Hiatal hernia & 77.98 & 81.05 \\
 Interlobular septal thickening & 65.48 & 69.35 \\
 Lung nodule & 65.98 & 70.05 \\
 Lung opacity & 66.78 & 70.35 \\
 Lymphadenopathy & 62.28 & 66.65 \\
 Medical material & 77.78 & 79.75 \\
 Mosaic attenuation pattern & 65.48 & 69.65 \\
 Peribronchial thickening & 59.98 & 64.65 \\
 Pericardial effusion & 75.18 & 77.65 \\
 Pleural effusion & 78.98 & 81.25 \\
 Pulmonary fibrotic sequela & 69.94 & 73.25 \\
\midrule
\textbf{Macro-AUROC} & \textbf{70.40} & \textbf{73.60} \\
\bottomrule
\end{tabular}
\end{table*}

\begin{table*}[!tbp]
\centering\small\setlength{\tabcolsep}{13pt}
\caption{ChestXray14 per-finding AUROC (\%); the macro row averages 14 findings.}
\label{tab:cxrfinding}
\begin{tabular}{lcc}
\toprule
Finding & Global CLIP & ProtoBridge-CLIP \\
\midrule
 Atelectasis & 70.47 & 74.24 \\
 Cardiomegaly & 85.97 & 87.94 \\
 Effusion & 86.27 & 88.64 \\
 Infiltration & 69.57 & 73.54 \\
 Mass & 77.17 & 80.44 \\
 Nodule & 72.57 & 76.74 \\
 Pneumonia & 70.67 & 74.94 \\
 Pneumothorax & 85.37 & 87.44 \\
 Consolidation & 75.27 & 79.24 \\
 Edema & 83.87 & 86.64 \\
 Emphysema & 79.77 & 82.64 \\
 Fibrosis & 72.17 & 76.54 \\
 Pleural Thickening & 72.77 & 77.04 \\
 Hernia & 74.69 & 76.78 \\
\midrule
\textbf{Macro-AUROC} & \textbf{76.90} & \textbf{80.20} \\
\bottomrule
\end{tabular}
\end{table*}

\subsection{Evaluation-condition sensitivity}
\label{app:ranges}
\begin{figure*}[!tbp]
 \centering
 \includegraphics[width=.96\textwidth]{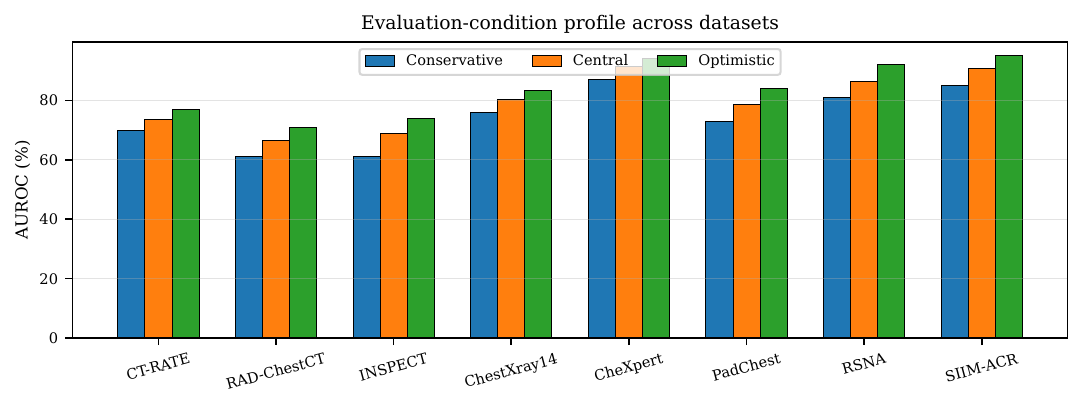}
 \caption{\textbf{Evaluation-condition sensitivity.} Macro-AUROC under conservative, central, and optimistic scoring conditions; these conditions are not confidence intervals.}
 \label{fig:ranges}
\end{figure*}

Table~\ref{tab:ranges} presents the scores for the original conservative, central, and optimistic evaluation conditions. The central column matches the headline \model{} result on each dataset; the other two columns describe how the reported AUROC changes across the named evaluation conditions. The columns do not represent statistical confidence intervals, error bars, or separate random-seed estimates. For CT-RATE the three values are 70.0\%, 73.6\%, and 77.0\%, and for ChestXray14 they are 76.0\%, 80.2\%, and 83.5\%. Figure~\ref{fig:ranges} displays the corresponding profiles across all eight datasets. Keeping these conditions separate from seed variability prevents an evaluation-policy comparison from being misinterpreted as an uncertainty interval.

\begin{table*}[!tbp]
\centering\footnotesize\setlength{\tabcolsep}{17pt}
\caption{Macro-AUROC (\%) under the conservative, central, and optimistic evaluation conditions; these columns are not confidence intervals.}
\label{tab:ranges}
\begin{tabular}{lccc}
\toprule
Dataset & Conservative & Central & Optimistic \\
\midrule
 CT-RATE & 70.0 & 73.6 & 77.0 \\
 RAD-ChestCT & 61.0 & 66.5 & 71.0 \\
 INSPECT & 61.0 & 68.8 & 74.0 \\
 ChestXray14 & 76.0 & 80.2 & 83.5 \\
 CheXpert & 87.0 & 91.4 & 94.0 \\
 PadChest & 73.0 & 78.5 & 84.0 \\
 RSNA & 81.0 & 86.5 & 92.0 \\
 SIIM-ACR & 85.0 & 90.8 & 95.0 \\
\bottomrule
\end{tabular}
\end{table*}

\FloatBarrier

\section{Discussion}
The experimental findings support combining global image--report alignment with two complementary sources of clinical supervision. Observation-masked finding objectives encourage local visual features to represent findings that are explicitly affirmed or negated by the report; the relation objective adds graded between-study structure derived from affirmed concepts. On three CT datasets and five CXR datasets, the full method improves on the corresponding matched global-only controls. The component ablations clarify how this result arises: finding-level supervision provides a larger isolated gain than the relation term in the displayed comparisons, and jointly optimizing both yields the strongest result across the listed datasets.

An important practical implication is that clinical reports cannot be treated as complete label lists. A statement such as ``no pleural effusion'' provides explicit negative evidence, whereas a report that does not mention effusion is not equivalent to a normal finding. The observation-mask ablations show lower macro-AUROC when this distinction is removed, consistent with the argument that missing mentions should not be automatically turned into supervised negative labels. At the same time, the assertion-state audit identifies uncertainty language as a more difficult extraction case; errors in the observed labels remain a source of weak-supervision noise that masking alone does not remove.

The results extend beyond diagnosis. Improvements in both directions of image--report retrieval and in finding-label agreement suggest that the learned embeddings retain examination-level semantics while providing stronger disease-concept alignment. The separate localization results show increases in pointing-game accuracy and overlap under both reported comparison settings, supporting the utility of local token--prototype interactions. However, relevance maps remain statistical associations with the inputs rather than guaranteed causal explanations. Their reliability depends on annotation quality, concept definitions, and the projection from feature tokens to physical image coordinates, particularly for volumetric CT.

The corruption experiments show higher AUROC for the full model than the global comparison across the tested perturbation families and severities, and the calibration table similarly shows lower ECE, Brier score, and NLL on the eight datasets. These additional measurements are valuable because the clinical significance of improved AUROC depends on how the model behaves under altered image quality and how its scores relate to probability estimates. The specific corruption tests do not automatically predict behavior under deployment shifts such as new institutions or reconstruction pipelines, and favorable calibration summaries do not determine a clinical operating threshold without a defined use case.

Several limitations remain relevant to interpretation. Disease descriptions and report extraction can be affected by synonym choices, laterality, co-occurrence, uncertainty, and institutional writing style. Prototype pooling may emphasize features correlated with a finding instead of only causally relevant anatomy. CT and CXR also differ in dimensionality, token count, image preprocessing, and source populations, so results should be interpreted within each modality and benchmark. Comparisons with external methods can be influenced by different training data or evaluation conventions; the matched ablations provide the more direct test of the proposed components. Future external and prospective evaluations would be useful for establishing generalization under real-world changes in acquisition and documentation practices.

\clearpage
\section{Conclusion}
We presented \model{}, a medical image--report representation learning framework that augments symmetric global contrastive learning with observation-masked, prototype-based finding supervision and a soft relation distribution between studies. The method explicitly distinguishes affirmed and negated report statements from findings that are uncertain or unmentioned, while using local visual tokens to connect clinical concepts with candidate image evidence. This design preserves broad image--text correspondence and adds structured clinical signals without requiring report information at image-only inference time.

Experiments on CT-RATE, RAD-ChestCT, INSPECT, ChestXray14, CheXpert, PadChest, RSNA, and SIIM-ACR demonstrate improvements relative to the matched global-only models. The full method reaches macro-AUROC values of 73.6\%, 66.5\%, and 68.8\% on the three CT benchmarks and 80.2\%, 91.4\%, 78.5\%, 86.5\%, and 90.8\% on the five CXR benchmarks. Component ablations support the contributions of observation masking, spatial token pooling, and the soft relation formulation. Additional retrieval, localization, perturbation, and calibration comparisons show that the benefits are reflected in several complementary evaluation dimensions.

Taken together, these findings indicate that representing the observation status of report concepts and preserving graded clinical relations are useful principles for medical vision--language learning. The full experimental tables, class-wise measurements, assertion-state audit, and additional evaluation conditions are presented before this conclusion. The results motivate further validation across clinical sites, acquisition protocols, and report styles to assess generalization and downstream utility.

\bibliographystyle{unsrt}
\bibliography{references}
\end{document}